\documentclass[10pt,twocolumn,letterpaper]{article}

\usepackage{iccv}
\usepackage{times}
\usepackage{epsfig}
\usepackage{graphicx}
\usepackage{amsmath}
\usepackage{amssymb}
\usepackage{tabularx, booktabs}
\usepackage{authblk}

\newcommand{\minisection}[1]{\vspace{1mm}\noindent{\bf #1}}
\def\mbf#1{\mathbf{#1}}

\usepackage[breaklinks=true,bookmarks=false]{hyperref}

\iccvfinalcopy

\begin{document}

\title{Temporal Tessellation: A Unified Approach for Video Analysis}

\author[1]{Dotan Kaufman}
\author[1]{Gil Levi}
\author[2,3]{Tal Hassner}
\author[1,4]{Lior Wolf}

\affil[1]{The Blavatnik School of Computer Science \\
Tel Aviv University\\ Israel  }
\affil[2]{Information Sciences Institute \\ USC \\CA\\ USA } 
\affil[3]{The Open University of Israel\\Israel } 
\affil[4]{Facebook AI Research}

\maketitle

\begin{abstract}
We present a general approach to video understanding, inspired by semantic transfer techniques that have been successfully used for 2D image analysis. Our method considers a video to be a 1D sequence of clips, each one associated with its own semantics. The nature of these semantics -- natural language captions or other labels -- depends on the task at hand. A test video is processed by forming correspondences between its clips and the clips of reference videos with known semantics, following which, reference semantics can be transferred to the test video. We describe two matching methods, both designed to ensure that (a) reference clips appear similar to test clips and (b), taken together, the semantics of the selected reference clips is consistent and maintains temporal coherence. We use our method for video captioning on the LSMDC'16 benchmark, video summarization on the SumMe and TVSum benchmarks,  Temporal Action Detection on the Thumos2014 benchmark, and sound prediction on the Greatest Hits benchmark.  Our method not only surpasses the state of the art, in four out of five benchmarks, but importantly, it is the only single method we know of that was successfully applied to such a diverse range of tasks.
\end{abstract}

\section{Introduction}
Despite decades of research, video understanding still challenges computer vision. The reasons for this are many, and include the hurdles of collecting, labeling and processing video data, which is typically much larger yet less abundant than images. Another reason is the inherent ambiguity of actions in videos which often defy attempts to attach dichotomic labels to video sequences~\cite{kliper2012action}

Rather than attempting to assign videos with single {\em action labels} (in the same way that 2D images are assigned object classes in, say, the ImageNet collection~\cite{russakovsky2015imagenet}) an increasing number of efforts focus on other representations for the semantics of videos. One popular approach assigns videos with natural language text annotations which describe the events taking place in the video~\cite{chen2011collecting,rohrbach15cvpr}. Systems are then designed to automatically predict these annotations. Others attach video sequences with numeric values indicating what parts of the video are more interesting or important~\cite{gygli2014creating}. Machine vision is then expected to determine the importance of each part of the video and summarize videos by keeping only their most important parts. 

\begin{figure}[t]
\centering
\includegraphics[width=0.48\textwidth,clip,trim = 0mm 0mm 0mm 0mm]{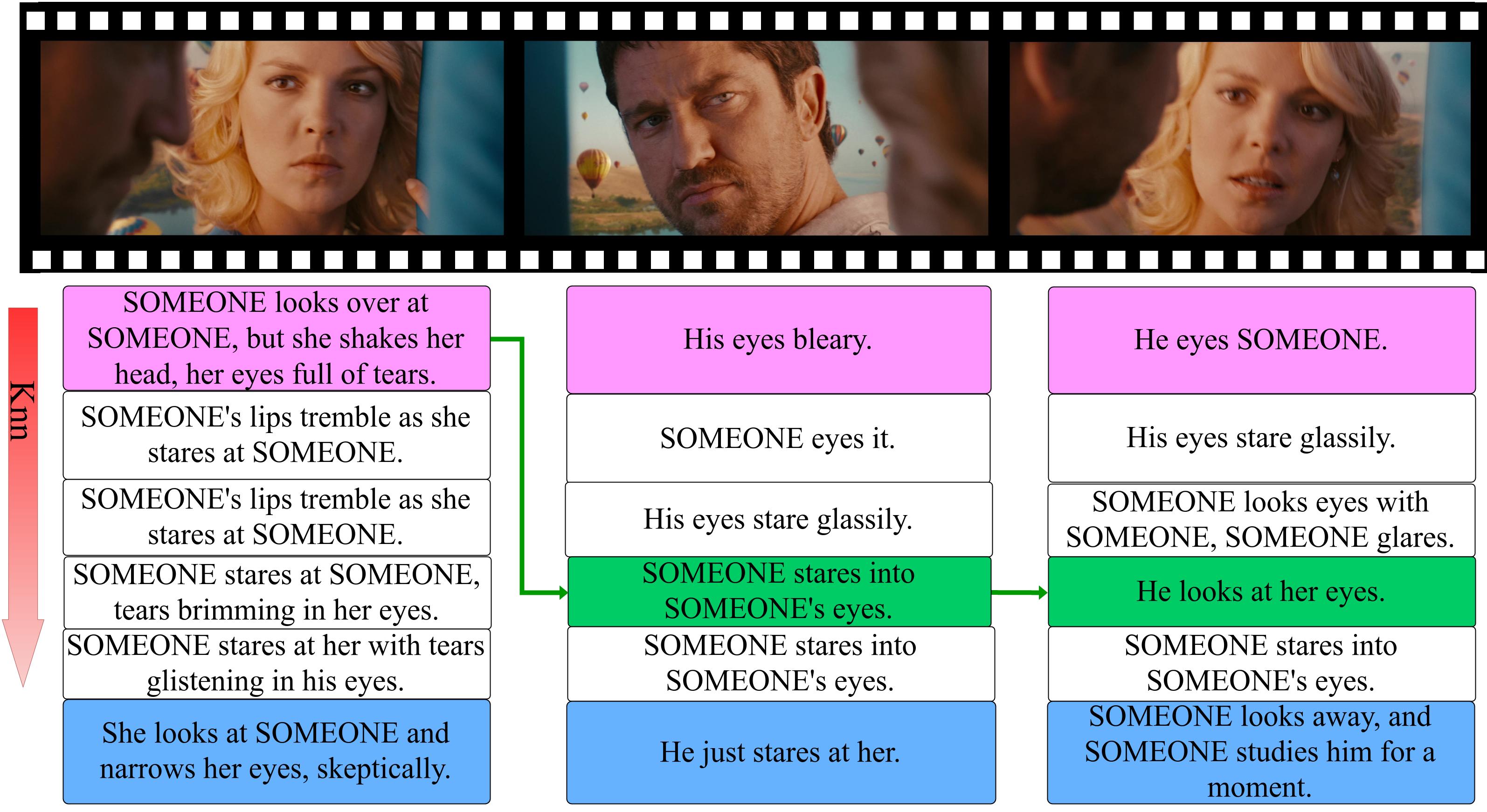}
\caption{{\bf Tessellation for temporal coherence}. For video captioning, given a query video (top), we seek reference video clips with similar semantics. Our tessellation ensures that the semantics assigned to the test clip are not only the most relevant (the five options for each clip) but also preserve temporal coherence (green path). Ground truth captions are provided in blue.}\label{fig:teaser}
\end{figure}

Although impressive progress was made on these and other video understanding problems, this progress was often made disjointedly: separate specialized systems were utilized that were tailored to obtain state of the art performance on different video understanding problems. Still lacking is a {\em unified} general approach to solving these different tasks. 

Our approach is inspired by recent 2D dense correspondence estimation methods (e.g.,~\cite{hassner2015dense,liu2011sift}). These methods were successfully shown to solve a variety of image understanding problems by transferring per-pixel semantics from reference images to query images. This general approach was effectively applied to a variety of tasks, including single view depth estimation, semantic segmentation and more. We take an analogous approach, applying similar techniques to 1D video sequences rather than 2D images. 

Specifically, image based methods combine local, per-pixel appearance similarity with global, spatial smoothness. We instead combine local, per-region appearance similarity with global semantics smoothness, or {\em temporal coherence}. Fig.~\ref{fig:teaser} offers an example of this, showing how temporal coherence improves the text captions assigned to a video.

Our contributions are as follows: {\bf (a)} We describe a novel method for matching test video clips to reference clips. References are assumed to be associated with semantics representing the task at hand. Therefore, by this matching we transfer semantics from reference to test videos. This process seeks to match clips which share similar appearances while maintaining semantic coherency between the assigned reference clips. {\bf (b)} We discuss two techniques for maintaining temporal coherency: the first uses unsupervised learning for this purpose whereas the second is supervised. 

Finally, {\bf (c)}, we show that our method is general by presenting state of the art results on three recent and challenging video understanding tasks, previously addressed separately: Video caption generation on the LSMDC'16 benchmark~\cite{lsmdc2015}, video summarization on the SumMe~\cite{gygli2014creating} and TVSum~\cite{song2015tvsum} benchmarks, and action detection on the THUMOS'14 benchmark~\cite{THUMOS14}. In addition, we report results comparable to the state of the art on the Greatest Hits benchmark~\cite{owens2016visually} for sound prediction from video.  Importantly, we will {\em publicly release our code and  models}.\footnote{See:~\url{github.com/dot27/temporal-tessellation}}

\section{Related work}\label{sec:related}
\minisection{Video annotation.} Significant progress was made in the relatively short time since work on video annotation / caption generation began. Early methods such as~\cite{aradhye2009video2text,huang2012multi,over2012trecvid,wei2010multimodal} attempted to cluster captions and videos and applied this for video retrieval. Others~\cite{guadarrama2013youtube2text,krishnamoorthy2013generating,thomason2014integrating} generated sentence representations by first identifying semantic video content (e.g., verb, noun, etc.) using classifiers tailored for particular objects and events. They then produce template based sentences. This approach, however, does not scale well, since it requires substantial efforts to provide suitable training data for the classifiers, as well as limits the possible sentences that the model can produce. 

More recently, and following the success of image annotation systems based on deep networks such as~\cite{donahue2015long,vinyals2015show}, similar techniques were applied to videos~\cite{donahue2015long,srivastava2015unsupervised,venugopalan2015sequence,yao2015describing}. Whereas image based methods used convolutional neural networks (CNN) for this purpose, application to video involve temporal data, which led to the use of recurrent neural networks (RNN), particularly short-term memory networks (LSTM)~\cite{hochreiter1997long}. We also use CNN and LSTM models but in fundamentally different ways, as we later explain in Sec.~\ref{sec:tessellation}. 

\minisection{Video summarization.} This task involves selecting the subset of a query video's frames which represents its most important content. Early methods developed for this purpose relied on manually specified cues for determining which parts of a video are important and should be retained. A few such examples include~\cite{chu2015video,potapov2014category,song2015tvsum,zhang1997integrated}. 

More recently, the focus shifted towards supervised learning  methods~\cite{gong2014diverse,gygli2014creating,gygli2015video,zhang2016summary}, which assume that training videos also provide manually specified labels indicating the importance of different video scenes. These methods sometimes use multiple individual-tailored decisions to choose video portions for the summary~\cite{gygli2014creating,gygli2015video} and often rely on the determinantal point process (DPP) in order to increase the diversity of selected video subsets~\cite{chao2015large,gong2014diverse,zhang2016summary}. 

Unlike video description, LSTM based methods were only considered for summarization very recently~\cite{zhang2016video}. Their use of LSTM is also very different from ours.

\minisection{Temporal action detection.} Early work on video action recognition relied on hand crafted space-time features~\cite{kliper2012motion,laptev2005space,wang2013action}. More recently, deep methods have been proposed~\cite{ji20133d,karpathy2014large,taylor2010convolutional}, many of which learn deep visual and motion features~\cite{li2016vlad3,simonyan2014two,tran2015learning,wang2015action}. Along with the development of stronger methods, larger and more challenging benchmarks were proposed~\cite{hassner2013critical, kliper2012action,kuehne2011hmdb,soomro2012ucf101}. Most datasets, however, used trimmed, temporally segmented videos, i.e: short clips which contain only a single action.

Recently, similar to the shift toward classification combined with localization in object recognition, some of the focus shifted toward more challenging and realistic scenarios of classifying untrimmed videos~\cite{caba2015activitynet,THUMOS14}. In these datasets, a given video can be up to a few minutes in length, different actions occur at different times in the video and in some parts of the video no clear action occurs. These datasets are also used for classification, i.e. determining the main action taking place in the video. A more challenging task, however, is the combination of classification with temporal detection: determining which action, if any, is taking place at each time interval in the video. 

In order to tackle temporal action detection in untrimmed videos, Yuan et al.~\cite{yuan2016temporal} encode visual features at different temporal resolutions followed by a classifier to obtain classification scores at different time scales. Escorcia et al~\cite{escorcia2016daps} focus instead on a fast method for obtaining action proposals from untrimmed videos, which later can be fed to an action classifier. Instead of using action classifiers, our method relies on matching against a gallery of temporally segmented action clips.

\section{Preliminaries}\label{sec:preliminaries}
Our approach assumes that test videos are partitioned into clips. It then matches each test clip with a reference ({\em training}) clip. Matching is performed with two goals in mind. First, at the clip level, we select reference clips which are visually similar to the input. Second, at the video level, we select a sequence of clips which best preserves the temporal semantic coherency. Taken in sequence, the order of selected, reference semantics should adhere to the temporal manner in which they appeared in the training videos. 

Following this step, the semantics associated with selected reference clips can be transferred to test clips. This allows us to reason about the test video using information from our reference. This approach is general, since it allows for different types of semantics to be stored and transferred from reference, training videos to the test videos. This can include, in particular, textual annotations, action labels, manual annotations of interesting frames and others. Thus, different semantics represent different video understanding problems which our method can be used to solve.

\subsection{Encoding video content}\label{sec:appearance}
We assume that training and test videos are partitioned into sequences of clips. A clip $\mbf{C}$ consists of a few consecutive frames $\mbf{I}_i, i\in 1..n$ where $n$ is the number of frames in the clip. Our tessellation approach is agnostic to the particular method chosen to represent these clips. Of course, The more robust and discriminative the representation, the better we expect our results to be. We, therefore, chose the following two step process, based on the recent state of the art video representations of~\cite{lev2015rnn}.

\minisection{Step 1: Representing a single frame.} Given a frame $\mbf{I}_i$ we use an off the shelf CNN to encode its appearance. We found the VGG-19 CNN to be well suited for this purpose. This network was recently proposed in~\cite{simonyan2014very} and used to obtain state of the art results on the ImageNet, large scale image recognition
benchmark (ILSVRC)~\cite{russakovsky2015imagenet}. In their work,~\cite{simonyan2014very} used the last layer of this network to predict ImageNet class labels, represented as one-hot encodings. We instead treat this network as a feature transform function $f:\mbf{I} \mapsto \boldsymbol{a}'$ which for image (frame) $\mathbf{I}$ returns the $4,096D$ response vector from the penultimate layer of the network. 

To provide robustness to local translations, we extract these features by oversampling: $\mathbf{I}$ is cropped ten times at different offsets around the center of the frame. These cropped frames are normalized by subtracting the mean value of each color channel and then fed to the network. Finally, the ten $4,096D$ response vectors returned by the network are pooled into a single vector by element-wise averaging. Principle component analysis (PCA) is further used to reduce the dimensionality of these features to $500D$, giving us the final, per frame representation $\mbf{a}\in\mathbb{R}^{500}$.

\minisection{Step 2: Representing multiple frames.} Once the frames are encoded, we pool them to obtain a representation for the entire clip. Pooling is performed by Recurrent Neural Network Fisher Vector (RNN-FV) encoding~\cite{lev2015rnn}. 

Specifically, We use their RNN, trained to predict the feature encoding of a future frame, $\mbf{a}_{i}$, given the encodings for its $k$ preceding frames, $(\mbf{a}_{i-k},...,\mbf{a}_{i-1})$. This RNN was trained on the training set from the Large Scale Movie Description Challenge~\cite{lsmdc2015}, containing roughly 100K videos. We apply the RNN-FV to the representations produced for all of the frames in the clip. The gradient of the last layer of this RNN is then taken as a 100,500D representation for the entire sequence of frames in $\mbf{C}$. We again use PCA for dimensionality reduction, this time mapping the features produced by the RNN-FV to 2,000D dimensions, resulting in our pooled representation $\mbf{A} \in \mathbb{R}^{2,000}$. We refer to~\cite{lev2015rnn} for more details about this process.

\subsection{Encoding semantics}\label{sec:semantics}
As previously mentioned, the nature of the semantics associated with a video depends on the task at hand. For tasks such as action detection and video summarization, for which the supervision signal is of low dimension, the semantic space of the labels has only a few bits of information per segment and is not discriminative enough between segments. In this case, we take the semantic space $\mbf{V}^S$ to be the same as the appearance space $\mbf{V}^A$ and take both to be the pooled representation $\mbf{A}$. 

\minisection{Textual semantics}  In video captioning, in which the text data provides a rich source of information, our method largely benefits from having a separate semantic representation that is based on the label data.  

We tested several representations for video semantics and chose the recent Fisher Vector of a Hybrid Gaussian-Laplacian Mixture Model (FV-HGLMM)~\cite{klein2015associating}, since it provided the best results in our initial cross-validation experiments.

Briefly, we assume a textual semantic representation, $\mbf{s}$ for a clip $\mbf{C}$, where $\mbf{s}$ is a string containing natural language words. We use word2vec~\cite{mikolov2013efficient} to map the sequence of words in $\mbf{s}$ to a vector of numbers, $(s_1,...,s_m)$, where $m$ is the number of words in $\mbf{s}$ and can be different for different clips. FV-HGLMM then maps this sequence of numbers to a vector $\mbf{S}\in\mathbb{R}^M$ of fixed dimensionality, $M$. 

FV-HGLMM is based on the well-known Fisher Vectors (FV)~\cite{perronnin2010improving,simonyan2013deep,sydorov2014deep}. The standard Gaussian Mixture Models (GMM) typically used to produce FV representations are replaced here with a Hybrid Gaussian-Laplacian Mixture Model which was shown in~\cite{klein2015associating} to be effective for image annotation. We refer to that paper for more details.

\subsection{The joint semantics video space (SVS)}\label{sec:svs}
Clip representations and their associated semantics are all mapped to the joint SVS. We aim to map the appearance of each clip and its assigned semantics to two neighboring points in the SVS. By doing so, given an {\em appearance} representation for a query clip, we can search for potential {\em semantic} assignments for this clip in our reference set using standard Euclidean distance. This property will later become important in Sec.~\ref{sec:distribution}. 

In practice, all clip appearance representations $\mbf{A}$ and their associated semantic representations $\mbf{S}$ are jointly mapped to the SVS using regularized Canonical Correlation Analysis (CCA)~\cite{regularized_cca} where the CCA mapping is trained using the ground truth semantics provided with each benchmark. In our experiments, the CCA regularization parameter is fixed to be a tenth of the largest eigenvalue of the cross domain covariance matrix computed by CCA. For each clip, CCA projects $\mbf{A}$ and $\mbf{S}$ (appearance and semantics, respectively) to $\mbf{V}^A$ and $\mbf{V}^S$. 

\begin{figure*}[t]
\centering
\begin{tabular}{cc}
\includegraphics[width=0.49\textwidth,clip,trim = 0mm 0mm 0mm 0mm]{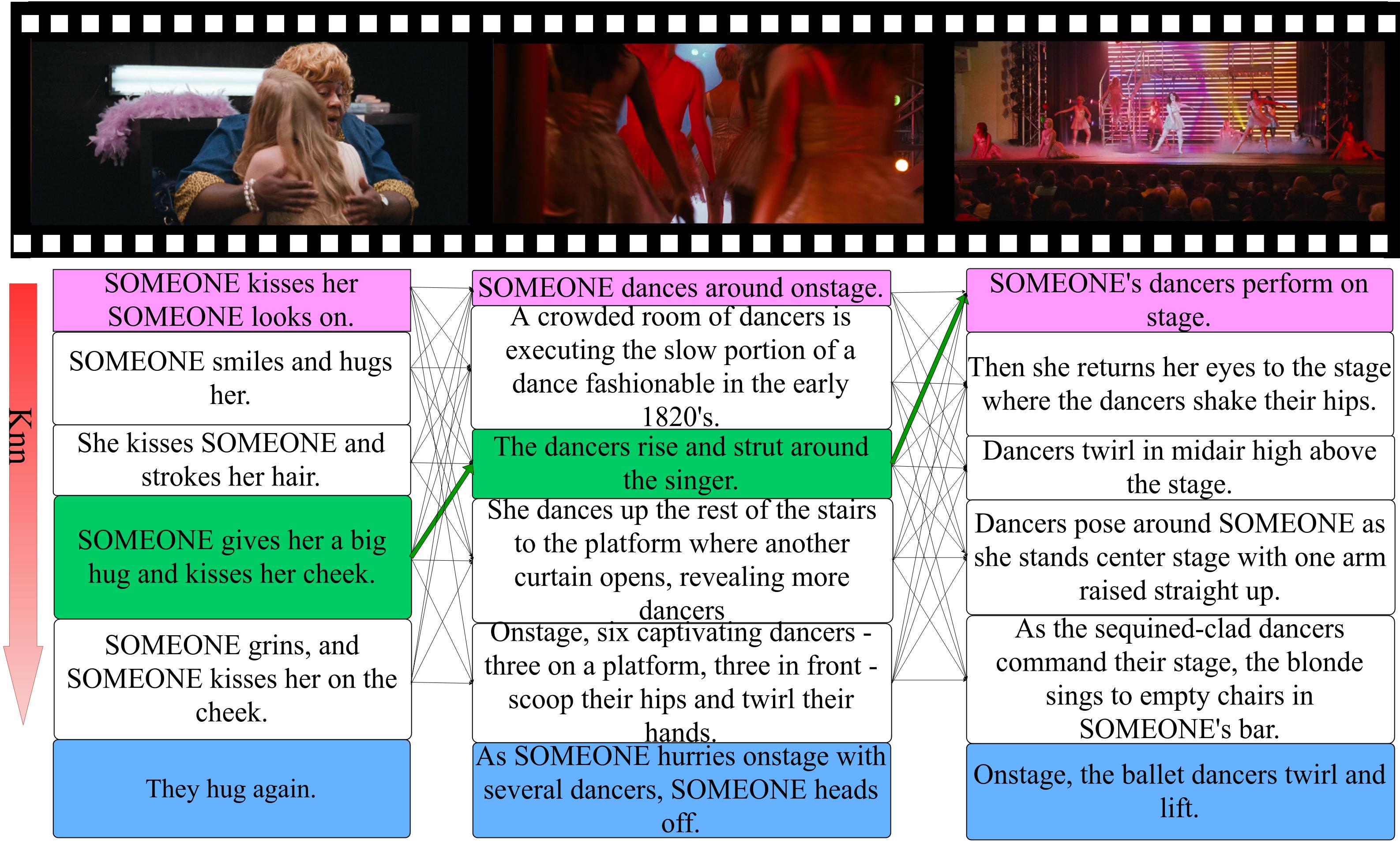}&
\includegraphics[width=0.49\textwidth,clip,trim = 0mm 0mm 0mm 0mm]{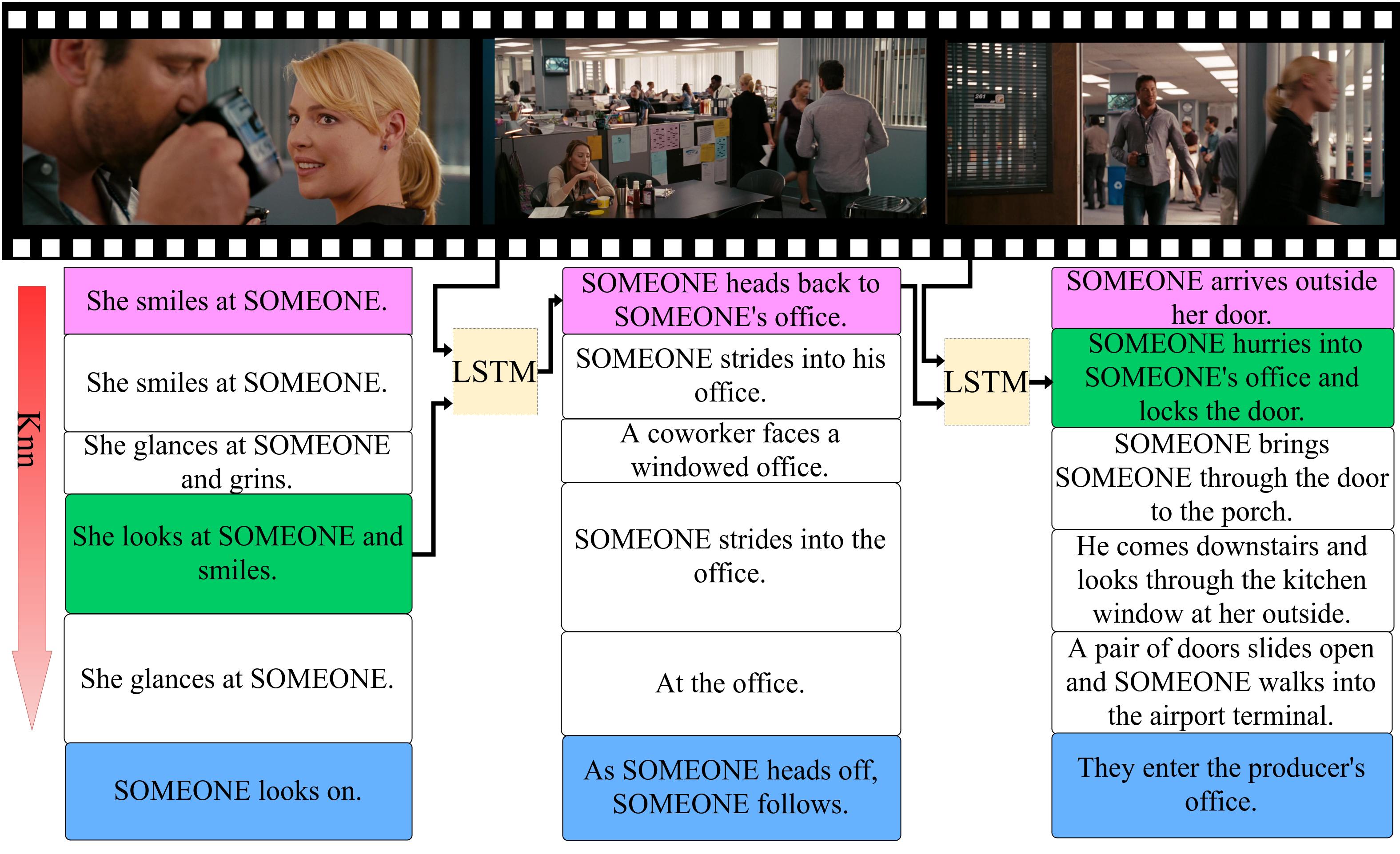}
\end{tabular}
\caption{{\bf Our two non-local tessellations}. {\bf Left:} Tessellation by restricted Viterbi. For a query video (top), our method finds visually similar videos and selects the clips that preserve temporal coherence using the Viterbi Method. The ground truth captions are shown in blue, the closest caption is shown in pink. Note that our method does not always select clips with the closest captions but the ones that best preserve temporal coherence. {\bf Right:} Tessellation by predicting the dynamics of semantics. Given a query video (top) and a previous clip selection, we use an LSTM to predict the most accurate semantics for the next clip.}\label{fig:methods}
\end{figure*}

\section{Tessellation}\label{sec:tessellation}
We assume a data set of training (reference) clips, $\mbf{V}^A_j$, and their associated semantics, $\mbf{V}^S_j$, represented as described in Sec.~\ref{sec:preliminaries}. Here, $j\in 1..N$ indexes the entire data set of $N$ clips. Since these clips may come from different videos, $j$ does not necessarily reflect temporal order. 

Given a test video, we process its clips following~\ref{sec:appearance} and~\ref{sec:svs}, obtaining a sequence of clip representations, $\mbf{U}^A_i$ in the SVS, where consecutive index values for $i\in M$, represent consecutive clips in a test video with $M$ clips. Our goal is to match each $\mbf{U}^A_i$ with a data set {\em semantic} representation $\mbf{V}^S_{j_i}$ while optimizing the following two requirements:
\begin{enumerate}
\item {\bf Semantics-appearance similarity.} The representation for the test clip {\em appearance} is similar to the representation of the selected {\em semantics}.
\item {\bf Temporal coherence.} The selected semantics are ordered similar to their occurrences in the training set.
\end{enumerate}
Drawing on the analogy to spatial correspondence estimation methods such as SIFT flow~\cite{liu2011sift}, the first requirement is a {\em data term} and the second is a {\em smoothness term}, albeit with two important distinctions: First, the data term matches test {\em appearances} to reference {\em semantics} directly, building on the joint embedding of semantics and appearances in the SVS. Second, we define the smoothness term in terms of associated semantics and not pixel coordinates. 

\subsection{Local Tessellation}

Given the sequence of appearance representations $\mathcal{U}=(\mbf{U}^A_1, ..., \mbf{U}^A_M)$ for the test sequence, we seek a corresponding set of reference semantics $\mathcal{V}=(\mbf{V}^S_{j_1}, ..., \mbf{V}^S_{j_M})$ (here, again, $j$ indexes the $N$ clips in the reference set). The local tessellation method employs only the semantics-appearance similarity. In other words, we associate each test clip $\mbf{U}^A_i$, with the following training clip:
\begin{equation}
\mathcal{V}_{j_i}^*=\arg \min_{\mathcal{V}_j} ||\mbf{U}^A_i - \mbf{V}^S_{j} || 
\end{equation}

\subsection{Tessellation Distribution} \label{sec:distribution}

We make the Markovian assumption that the semantics assigned to input clip $i$, only depend on the appearance of clip $i$ and the semantics assigned to its preceding clip, $i-1$. This gives the standard factorization of the joint distribution for the clip appearances and their selected semantics:
\begin{align}
P(\mathcal{V},\mathcal{U}) =& P(\mbf{V}^S_{j_1})P(\mbf{U}^A_1|\mbf{V}^S_{j_1})\times\\
& \prod_{i=2}^M{P(\mbf{V}^S_{j_i}|\mbf{V}^S_{j_{i-1}})P(\mbf{U}^A_i|\mbf{V}^S_{j_i})}.\nonumber
\end{align}
We set the priors $P(\mbf{V}^S_{j_1})$ to be the uniform distribution. Due to our mapping of both appearances and semantics to the joint SVS, we can define both posterior probabilities simply using the L2-norm of these representations:
\begin{align}
P(\mbf{U}^A_i|\mbf{V}^S_j) &\propto \exp{(-||\mbf{U}^A_i-\mbf{V}^S_j||^2)}\label{eq:dataterm}\\
P(\mbf{V}^S_{j_i}|\mbf{V}^S_{j_{i-1}}) &\propto \exp{(-||\mbf{V}^S_{j_i}-\mbf{V}^S_{j_{i-1}}||^2)}\label{eq:smoothnes}
\end{align}

Ostensibly, We can now apply the standard Viterbi method~\cite{rabiner1989tutorial} to obtain a sequence $\mathcal{V}$ which maximizes this probability. In practice, we used a slightly modified version of this method, and, when possible, a novel method designed to better exploit our training data to predict database matches. These are explained below.

\subsection{Restricted Viterbi Method.} \label{sec:smooth}
Given the test clip appearance representations $\mathcal{U}$, the Viterbi method provides an assignment $\mathcal{V}^*$ such that,
\begin{equation}
\mathcal{V}^*=\arg \max_{\mathcal{V}} P(\mathcal{V},\mathcal{U}).
\end{equation}
We found that in practice $P(\mbf{U}^A_i|\mbf{V}^S_j)$ is a long-tail distribution, with only a few dataset elements $\mbf{V}^S_j$ near enough to any $\mbf{U}^A_i$ for their probability to be more than near-zero. We, therefore, restrict the Viterbi method in two ways. First, we consider only the $r'=5$ nearest neighboring database semantics features. Second, we apply a threshold on the probability of our data term, Eq.~(\ref{eq:dataterm}), and do not consider semantics $\mbf{V}^S_j$ falling below this threshold, except for the first nearest neighbor. Therefore, the number of available assignments for each clip is $1 \leq r\leq 5$. This process is illustrated in Figure ~\ref{fig:methods} (left).

\begin{table*}[t]
\centering
\begin{tabular}{lcc|ccccc}
\toprule
Method	                               &       	 CIDEr-D	&       	 BLEU-4 	 &       	 BLEU-1 	 &       	 BLEU-2 	 &       	 BLEU-3 	 &       	 METEOR 	 &       	 ROUGE 	\\ \hline
BUPT CIST AI lab$^*$ 	                     &       	 .072	   &       	 .005 	   &       	 .151 	   &       	 .047 	   &       	 .013 	   &       	 {\bf .075}&       	 .152 	\\
IIT Kanpur$^*$  	                          &       	 .042	   &       	 .004 	   &       	 .116 	   &       	 .003 	   &       	 .011 	   &       	 .070 	   &       	 .138 	\\
Aalto University$^*$ 	                     &       	 .037	   &       	 .002 	   &       	 .007 	   &       	 .001 	   &       	 .005 	   &       	 .033 	   &       	 .069 	\\ \hline
Shetty and Laaksonen~\cite{shetty2015video}&       	 .044	   &       	 .003 	   &       	 .119 	   &       	 .024 	   &       	 .007 	   &       	 .046 	   &       	 .108 	\\
Yu et al~\cite{yu2016video} 	              &       	 .082	   &       	 .007 	   &       	 .157 	   &       	 .049 	   &       	 {\bf .017}&       	 .070 	   &       	 .149 	\\
S2VT~\cite{venugopalan2015sequence} 	      &       	 .088	   &       	 .007 	   &       	 {\bf .162}&       	 {\bf.051}&       	 {\bf .017}&       	 .070 	   &       	 {\bf .157} 	\\ \hline
Appearance Matching & .042 & .003 & .118 & .026 & .008 & .046 & .110 \\
Local Tessellation 	                        &       	 .098	   &       	 .007 	   &       	.144  	   &       	 .042 	   &       	 .016 	   &       	 .056 	   &       	 .130 	\\
Unsupervised Tessellation 	                &       	.102	    &       	.007 	    &       	.146 	    &       	.043 	    &       	.016 	    &       	.055 	    &       	.137 	\\
Supervised Tessellation 	                  &       	{\bf .109}&       	{\bf .008}&       	.151 	    &       	.044 	    &       	{\bf .017}&       	.057 	    &       	.135 	\\
\bottomrule
\end{tabular}
\caption{{\em Video annotation results on the LSMDC'16 challenge~\cite{lsmdc2015}.} CIDEr-D and BLEU-4  values were found to be the most correlated with human annotations in~\cite{Rohrbach2016Large, vedantam2015cider}. Our results on these metrics far outperform others. * Denotes results which appear in the online challenge result board, but were never published. They are included here as reference.}
\label{tab:AnnotationResults}
\end{table*}

\subsection{Predicting the Dynamics of Semantics}\label{sec:predict}
The Viterbi method of Sec.~\ref{sec:smooth} is efficient and requires only unsupervised training. Its use of the smoothness term of Eq.~(\ref{eq:dataterm}), however, results in potentially constant semantic assignments, where for any $j_i$, $\mbf{V}^S_{j_i}$ can equal $\mbf{V}^S_{j_{i-1}}$. 

In cases where reference clips are abundant and come from continuous video sources, we provide a more effective method of ensuring smoothness. This is done by supervised learning of how the semantic labels associated with video clips change over time, and by using that to predict the assignment $\mbf{V}^S_{j_i}$ for $\mbf{U}^A_i$.

Our process is illustrated in Fig.~\ref{fig:methods} (right). We train an LSTM RNN~\cite{hochreiter1997long} on the semantic and appearance representations of the training set video clips. We use this network as a function:
\begin{gather}
g(\mbf{V}^S_{0},\mbf{V}^S_{1},...,\mbf{V}^S_{i-1},\mbf{U}^A_{1},...,\mbf{U}^A_{i-1},\mbf{U}^A_{i})=
\mbf{V}^S_{i},\nonumber\\
\mbf{V}^S_{0} = \mbf{0},
\end{gather}
which predicts the semantic representation $\mbf{V}^S_{i}$ for the clip at time $i$ given the semantic representation, $\mbf{V}^S_{i-1}$, assigned to the preceding clip and the appearance of the current clip, $\mbf{U}^A_{i}$. The labeled examples used to train $g$ are taken from the training set, following the processing described in Sec.~\ref{sec:semantics} and~\ref{sec:svs} in order to produce 2,000D post-CCA vectors. Each pair of previous ground truth semantics and current clip appearance in the training data provides one sample for training the LSTM. We employ two hidden layers, each with $1,000$ LSTM cells. The output, which predicts the semantics of the next clip, is also 2,000D.

Given a test video, we begin by processing it as in Sec.~\ref{sec:smooth}. In particular, for each of its clip representations $\mbf{U}^A_i$, we select $r\leq 5$ nearest neighboring semantics from the training set. At each time step $i$, we feed the clip and its assigned semantics from the preceding clip at time $i-1$ to our LSTM predictor $g$. We thus obtain an estimate for the semantics we expect to see at time $i$, $\hat{\mbf{V}}^S_i$. 

Of course, the predicted vector $\hat{\mbf{V}}^S_i$ cannot necessarily be interpreted as a semantic label: not all points in the SVS have semantic interpretations (in fact, many instead represent appearance information and more do not represent anything at all). We thus choose a representation ${\mbf{V}}^S_{j_i}$ out of the $r$ selected for this clip, such that $||\hat{\mbf{V}}^S_i - {\mbf{V}}^S_{j_i}||^2$ is smallest.

\section{Experiments}\label{sec:results}

We apply our method to four separate video understanding tasks: video annotation, video summarization, temporal action detection, and sound prediction. Importantly, previous work was separately tailored to each of these tasks; we are unaware of any previously proposed single method  which reported results on such a diverse range of video understanding problems. Contrary to the others, our method was applied to all of these tasks similarly. 

\subsection{Video Annotation}
In our annotation experiments, we used the movie annotation benchmark defined by the 2016 Large Scale Movie Description and Understanding Challenge (LSMDC16)~\cite{lsmdc2015}. LSMDC16 presents a unified version of the recently published large-scale movie datasets, M-VAD~\cite{AtorabiMVAD2015} and MPII-MD~\cite{rohrbach15cvpr}. The joint dataset contains 188 movies, divided to short (4-20 seconds) video clips with associated sentence descriptions. A total of 200 movies are available, from which 153, 12 and 17 are used for training, validation, and testing, respectively (20 more were used for blind tests, not performed here).

\begin{figure}[t]
\includegraphics[width=1.0\linewidth,clip,trim = 0mm 215mm 0mm 0mm]{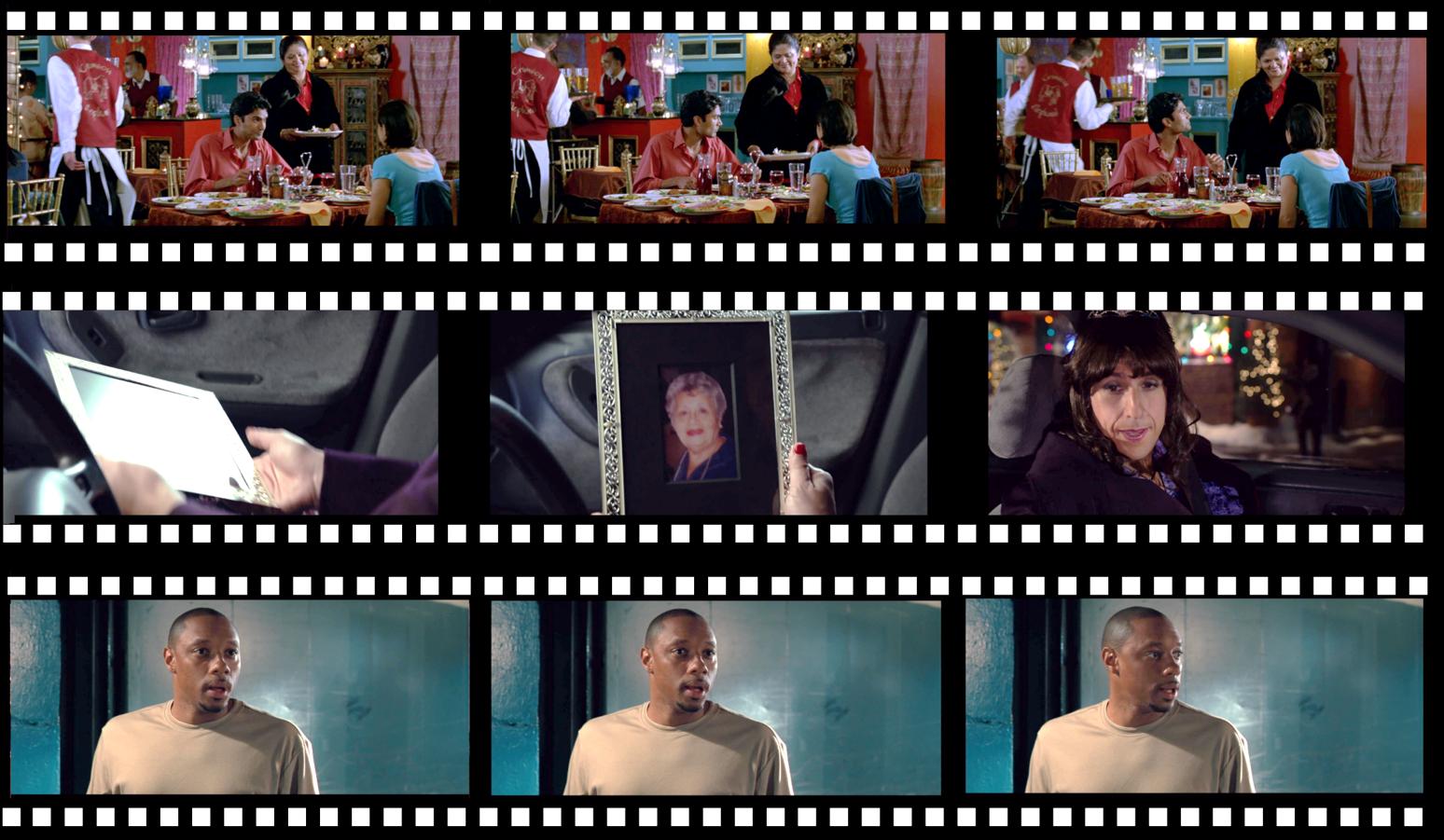}
GT: SOMEONE serves SOMEONE and SOMEONE. \\
ST: Now at a restaurant a waitress serves drinks. \\
\begin{footnotesize}~\\ \end{footnotesize}
\includegraphics[width=1.0\linewidth,clip,trim = 0mm 109mm 0mm 108mm]{Figure31.jpg}
GT: Then reaches in her bag and takes out a framed photo of a silver-haired woman.\\
ST: He spots a framed photo with SOMEONE in it.\\
\begin{footnotesize}~\\ \end{footnotesize}
\includegraphics[width=1.0\linewidth,clip,trim = 0mm 0mm 0mm 216mm]{Figure31.jpg}
GT: SOMEONE shifts his confused stare.\\
ST: He shifts his gaze then nods. 
\vspace{1mm}
\caption{{\bf Qualitative video captioning results}. Three caption assignments from the test set of the LSMDC16 benchmark. The Ground Truth captioning is provided along with the result of the Supervised Tessellation (ST) method.} 
\label{fig:qualitativecaption}
\end{figure}

Table~\ref{tab:AnnotationResults} present annotation results. We focus primarily on the CIDEr-D~\cite{vedantam2015cider} and the BLEU-4~\cite{papineni2002bleu} measures, since they are the only ones that are known to be well correlated with human perception~\cite{Rohrbach2016Large,vedantam2015cider}. Other metrics are provided here for completeness. These measures are: BLEU1--3~\cite{papineni2002bleu}, METEOR~\cite{denkowski2014meteor}, and ROUGE-L~\cite{lin2004automatic}. We compare our method with several published and unpublished systems. The results include the following three variants of our pipeline.

\minisection{Local tessellation.} Our baseline system uses per-clip nearest neighbor matching in the SVS in order to choose reference semantics. We match each test clip with its closest semantics in the SVS. From Tab.~\ref{tab:AnnotationResults}, we see that this method already outperforms previous State-of-the-Art. As reference, we provide the performance of a similar method which matches clips in appearance space ({\em Appearance matching}). The substantial gap between the two underscores the importance of our semantics-appearance similarity matching.

\minisection{Unsupervised tessellation.}
The graph-based method for considering temporal coherence, as presented in Sec.~\ref{sec:smooth} is able to provide a slight improvement in results in comparison to the local method (Tab.~\ref{tab:AnnotationResults}).

\minisection{Supervised tessellation.} The LSTM based model described in Sec.~\ref{sec:predict}, employing 2,000 LSTM units split between two layers. 

This method achieved the overall best performance on both CIDEr-D and BLEU-4, the metrics known to be most correlated with human perception~\cite{Rohrbach2016Large, vedantam2015cider}, outperforming previous state of the art with a gain of 23\% on CIDEr-D. Qualitative results are provided in Fig.~\ref{fig:qualitativecaption}.

\subsection{Video Summarization}

\begin{figure*}[t]
{\centering
\includegraphics[width=0.98\linewidth,clip,trim = 0mm 0mm 0mm 0mm]{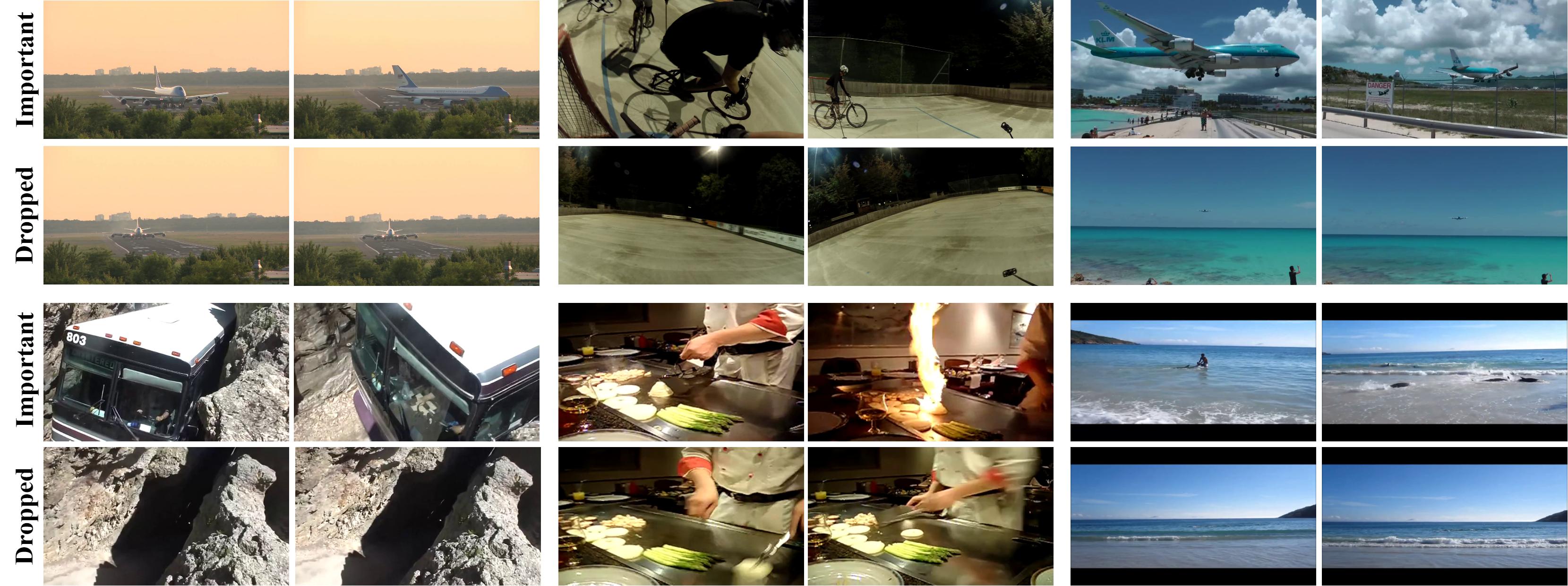}}
\caption{{\bf Sample video summarization results}. Sample frames from six videos out of the SumMe benchmark. Each group of four frames contains two frames (top rows) from short segments that were deemed important by the unsupervised tessellation method and two (bottom rows) that were dropped out of our summaries.}\label{fig:summary}
\end{figure*}

Video summarization performance is evaluated on the SumMe~\cite{gygli2014creating} and TVSum~\cite{song2015tvsum} benchmarks.
These benchmarks consist of 25 and 50 raw user videos, each depicting a certain event. The video frames are hand labeled with an importance score ranging from $0$ (redundant) and $1$ (vital) in SumMe and from $1$ (redundant) and $5$ (vital) in TVSum. The videos are about 1-5 minutes in length and the task is to produce a summary in the form of selected frames which is up-to 15\% of the given video's length. Sample frames are shown in Fig.~\ref{fig:summary}. The evaluation metric is the average f-measure of the predicted summary with the ground truth annotations. We follow~\cite{gygli2015video,zhang2016video} in evaluating with multiple user annotations.

Similar to video annotation, our approach is to transfer the semantics (represented here by frame importance values) from the gallery to the tessellated video. Our method operates without incorporating additional computational steps, such as optimizing the selected set using the determinantal point process~\cite{kulesza2012determinantal}, commonly used for such applications~\cite{chao2015large,gong2014diverse,zhang2016summary}.  

Table~\ref{tab:SummaryResults} compares our performance with several recent reports on the same benchmarks. We again provide results for all three variants of our system. This time, the local and the supervised tessellation methods are both outperformed by previous work on SumMe but not on TVSum. Our unsupervised tessellation outperforms other tessellation methods as well as the state of the art on the summarization benchmarks by substantial margins. 

We believe that unsupervised tessellation worked better than supervised because the available training examples were much fewer than required for the more powerful but data hungry LSTM. Specifically, for each benchmark we used only the labels from the same dataset, without leveraging other summarization datasets for this purpose. Doing so, using, e.g., the Open Video Project and YouTube dataset~\cite{de2011vsumm}, is left for future work.   

\begin{table}[t]
\centering
\begin{tabular}{lccc}
\toprule
Method$$ &  SumMe &  TVSum \\ \hline
Khosla et al.~\cite{khosla2013large} $\dagger$ $\ddagger$ & -- & 36.0 \\
Zhao et al.~\cite{zhao2014quasi} $\dagger$ $\ddagger$ & -- & 46.0 \\
Song et al.~\cite{song2015tvsum} $\dagger$  & -- & 50.0 \\
Gygli et. al~\cite{gygli2015video} & 39.7 & -- \\ 
Long Short-Term
Memory~\cite{zhang2016video} & 39.8  & 54.7 \\ 
Summary Transfer~\cite{zhang2016summary} & 40.9 & -- \\ \hline
Local Tessellation & 33.8 & 60.9 \\
Unsupervised Tessellation & \textbf{41.4} & \textbf{64.1} \\
Supervised Tessellation & 37.2 & 63.4 \\
\bottomrule
\end{tabular}
\caption{{\em Video summarization results on the SumMe~\cite{gygli2014creating} and TVSum~\cite{song2015tvsum} benchmarks.} Shown are the average f-measures. Our unsupervised tessellation method outperforms previous methods by a substantial margin.  $\dagger$ - unsupervised , $\ddagger$ - taken from~\cite{zhang2016video}}
\label{tab:SummaryResults}
\end{table}

\begin{figure*}[t]
{\centering
\includegraphics[width=0.98\linewidth,clip,trim = 0mm 0mm 0mm 0mm]{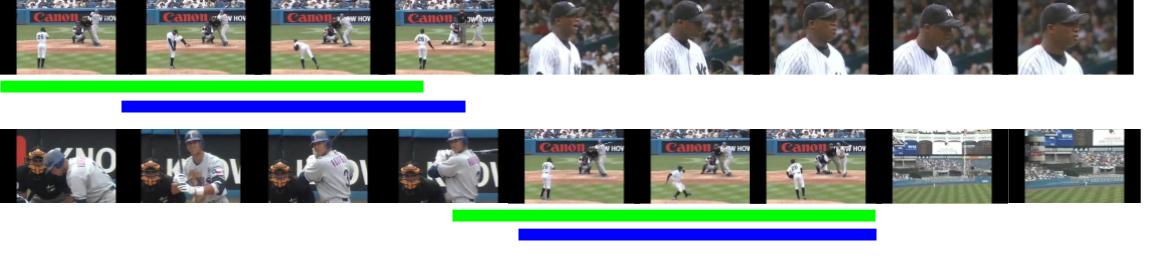}}
\caption{{\bf Sample action detection results}. Detection results for the 'Baseball pitch' class. The predicted intervals by the supervised tessellation method are shown in green, the ground truth in blue.}\label{fig:action}
\end{figure*}

\subsection{Temporal Action Detection}
We evaluate our method on the task of action detection, using the THUMOS'14~\cite{THUMOS14} benchmark for this purpose.

This one of the most recent and, to our knowledge, most challenging benchmark released for this task. THUMOS'14 consists of a training set of 13,320 temporally trimmed videos from the action classes of the UCF 101 dataset~\cite{soomro2012ucf101}, a validation set of 1,010 temporally untrimmed videos with temporal action annotations, a background set with 2,500 relevant videos guaranteed to not include any instance of the 101 actions and finally a test set with 1,574 temporally untrimmed videos. In the temporal action detection benchmark, for every action class out of a subset of 20 actions, the task is to predict both the presence of the action in a given video and its temporal interval, i.e., the start and end times of its detected instances.

For each action, the detected intervals are compared against ground-truth intervals using the Intersection over Union (IoU) similarity measure. Denoting the predicted intervals by $R_p$ and the ground truth intervals by $R_{gt}$, the IoU similarity is computed as 
$IoU = \frac{R_p \cap R_{gt}}{R_p \cup R_{gt}}$.

A predicted action interval is considered as true positive, if its IoT measure is above a predefined threshold and false positive otherwise. Ground truth annotations with no matching predictions are also counted as false positives. The Average Precision (AP) for each of the 20 classes is then computed and the mean Average Precision (mAP) serves as an overall performance measure. The process repeats for different IoT thresholds ranging from 0.1 to 0.5. 

Our approach to detection is to tessellate a given untrimmed video with short atomic clips from the UCF dataset~\cite{soomro2012ucf101}. With the resulting tessellation, we can determine which action occurred at each time in the video. Detection results on one sample video are shown in Fig.~\ref{fig:action}.

Tab.~\ref{tab:ActionResults} lists the results of the three variants of our framework along with previous results presented on the benchmark. As evident from the table, our tessellation method outperforms the state of the art by a large margin, where the supervised tessellation achieves the best results among the three variants of our method. Unsurprisingly, incorporating context in the tessellation dramatically improves performance, as can be seen from the relatively inferior local tessellation results.  

\begin{table}[t]
\centering
{
\begin{tabular}{lccccc} 
\toprule
Method$$ & 0.1  & 0.2 & 0.3 & 0.4 & 0.5 \\ \hline
Wang et al.~\cite{wang2014action} & 18.2 & 17.0 & 14.0 & 11.7 & 8.3 \\
Oneata et al.~\cite{oneata2014lear} & 36.6 & 33.6 & 27.0 & 20.8 & 14.4 \\
Heilbron et al.~\cite{caba2016fast} &-- & --& --& --& 13.5\\
Escorcia et al.~\cite{escorcia2016daps} & --&-- &-- &-- & 13.9\\
Richard and Gall~\cite{richard2016temporal} & 39.7 & 35.7 & 30.0 & 23.2 &  15.2\\
Shou et al.~\cite{shou2016temporal} & 47.7  & 43.5 & 36.3 & 28.7 & 19.0 \\
Yeung et al.~\cite{yeung2016end} & 48.9 & 44.0 & 36.0 & 26.4 & 17.1 \\
Yuan et al.~\cite{yuan2016temporal} & 51.4 & 42.6 & 33.6 & 26.1 &  18.8\\ \hline

Local tessellation & 56.4 & 51.2 &  43.8 &  32.5 & 20.7  \\
Unsupervised t. & 57.9 & 54.2 & 47.3 & 35.2  & 22.4 \\
Supervised t. & {\bf 61.1  } & {\bf 56.8} & {\bf 49.3} & {\bf 36.5}& {\bf 23.3} \\
\bottomrule
\end{tabular}
}
\caption{{\em Temporal Action Detection results on the THUMOS'14~\cite{THUMOS14} benchmark}. Shown are the mAP of the various methods for different IoT thresholds. Our proposed framework outperforms previous State-of-the-Art methods by a large margin. The supervised tessellation obtains the best results.}
\label{tab:ActionResults}
\end{table}

\subsection{Predicting Sounds from Video}
Though we show our method to obtain state of the art results on a number of tasks, at least in one case, a method tailored for a specific video analysis task performs better than our own. Specifically, we test the capability of our method to predict sound from video using the Greatest Hits dataset~\cite{owens2016visually}. This dataset consists of 977
videos of humans probing different environments with a drumstick: hitting, scratching, and
poking different objects in different scenes. Each
video, on average, contains 48 actions and lasts 35 seconds. In~\cite{owens2016visually}, a CNN followed by an LSTM was used to predict sounds for each video. Following their protocol, we consider only the video segments  centered on the audio amplitude peaks. We employ the published sound features that are available for 15 frame intervals around each audio peak, which we take to be our clip size. Each clip $\mbf{C}$ is therefore associated with the RNN-FV pooled VGG-19 CNN visual representation, as presented in Sec.~\ref{sec:appearance}, and with a vector $\mbf{a} \in \mathbb{R}^{1,890}$ of sound features that is simply the concatenation of these 15 sound features. 

Matching is performed in a SVS that is constructed from the visual representation and the matching sound features. We predict sound features for {\em hit events} by applying tessellation and returning the sound feature vector $\mbf{a}$ associated with the selected reference clips. 

There are two criteria that are used for evaluating the results: Loudness and Centroid. In both cases both the MSE scores and correlations are reported. Loudness is taken to be the maximum energy
(L2 norm) of the compressed subband envelopes over all
timesteps. Centroid is measured by taking the center of mass of
the frequency channels for a one-frame window around the center of the impact. 

Our results are reported in Tab.~\ref{tab:freeman}. The importance of the semantic space as can be observed from the gap between the appearance only matching to the Local Tessellation method. Leveraging our supervised and unsupervised tessellation methods improves the results even further.  In three out of four criteria the supervised tessellation seems preferable to the unsupervised one in this benchmark.

\begin{table}[t]
\centering
{
\begin{tabular}{lcccc} 
\toprule
Method$$ & \multicolumn{2}{c}{Loudness}  &    \multicolumn{2}{c}{Centroid}  \\
 & Err & $r$ & Err & $r$\\ \hline
 Full system of~\cite{owens2016visually} &{\bf 0.21}  &{\bf 0.44}   &{\bf 3.85}   &{ 0.47}    \\ \hline
Appearance matching &0.35 &0.18 &6.09 &0.36 \\
Local tessellation & 0.27 &0.32  &4.83  &0.47    \\
Unsupervised tessellation &0.26  &0.33   &4.76  &{\bf 0.48}\\
Supervised tessellation &0.24  &0.35   &4.44   &0.46    \\

\bottomrule
\end{tabular}
}
\caption{{\em Greatest Hits benchmark results}. Shown are the MSE and the correlation coefficient for two different success criteria.}
\label{tab:freeman}
\end{table}

\section{Conclusions}
We present a general approach to understanding and analyzing  videos. Our design transfers per-clip video semantics from reference, training videos to novel test videos. Three alternative methods are proposed for this transfer: local tessellation, which uses no context, unsupervised tessellation which uses dynamic programming to apply temporal, semantic coherency, and supervised tessellation which employs LSTM to predict future semantics. We show that these methods, coupled with a recent video representation technique, provide state of the art results on three very different video analysis domains: video annotation, video summarization, and action detection and near state of the art on a fourth application, sound prediction from video. Our method is unique in being first to obtain state of the art results on such different video understanding tasks, outperforming methods tailored for these applications.  

\section*{Acknowledgments}
This research is supported by the Intel Collaborative Research Institute for Computational Intelligence (ICRI-CI).

{\small
\bibliographystyle{ieee}

\begin{thebibliography}{10}\itemsep=-1pt

\bibitem{aradhye2009video2text}
H.~Aradhye, G.~Toderici, and J.~Yagnik.
\newblock Video2text: Learning to annotate video content.
\newblock In {\em Int. Conf. on Data Mining Workshops}, pages 144--151. IEEE,
  2009.

\bibitem{caba2016fast}
F.~Caba~Heilbron, J.~Carlos~Niebles, and B.~Ghanem.
\newblock Fast temporal activity proposals for efficient detection of human
  actions in untrimmed videos.
\newblock In {\em Proceedings of the IEEE Conference on Computer Vision and
  Pattern Recognition}, pages 1914--1923, 2016.

\bibitem{chao2015large}
W.-L. Chao, B.~Gong, K.~Grauman, and F.~Sha.
\newblock Large-margin determinantal point processes.
\newblock UAI, 2015.

\bibitem{chen2011collecting}
D.~L. Chen and W.~B. Dolan.
\newblock Collecting highly parallel data for paraphrase evaluation.
\newblock In {\em Proc. Annual Meeting of the Association for Computational
  Linguistics: Human Language Technologies}, pages 190--200. Association for
  Computational Linguistics, 2011.

\bibitem{chu2015video}
W.-S. Chu, Y.~Song, and A.~Jaimes.
\newblock Video co-summarization: Video summarization by visual co-occurrence.
\newblock In {\em Proc. Conf. Comput. Vision Pattern Recognition}, pages
  3584--3592, 2015.

\bibitem{de2011vsumm}
S.~E.~F. De~Avila, A.~P.~B. Lopes, A.~da~Luz, and A.~de~Albuquerque~Ara{\'u}jo.
\newblock Vsumm: A mechanism designed to produce static video summaries and a
  novel evaluation method.
\newblock {\em Pattern Recognition Letters}, 32(1):56--68, 2011.

\bibitem{denkowski2014meteor}
M.~Denkowski and A.~Lavie.
\newblock Meteor universal: Language specific translation evaluation for any
  target language.
\newblock In {\em In Proceedings of the Ninth Workshop on Statistical Machine
  Translation}. Citeseer, 2014.

\bibitem{donahue2015long}
J.~Donahue, L.~Anne~Hendricks, S.~Guadarrama, M.~Rohrbach, S.~Venugopalan,
  K.~Saenko, and T.~Darrell.
\newblock Long-term recurrent convolutional networks for visual recognition and
  description.
\newblock In {\em Proc. Conf. Comput. Vision Pattern Recognition}, pages
  2625--2634, 2015.

\bibitem{escorcia2016daps}
V.~Escorcia, F.~C. Heilbron, J.~C. Niebles, and B.~Ghanem.
\newblock Daps: Deep action proposals for action understanding.
\newblock In {\em European Conf. Comput. Vision}, pages 768--784. Springer,
  2016.

\bibitem{caba2015activitynet}
B.~G. Fabian Caba~Heilbron, Victor~Escorcia and J.~C. Niebles.
\newblock Activitynet: A large-scale video benchmark for human activity
  understanding.
\newblock In {\em Proc. Conf. Comput. Vision Pattern Recognition}, pages
  961--970, 2015.

\bibitem{gong2014diverse}
B.~Gong, W.-L. Chao, K.~Grauman, and F.~Sha.
\newblock Diverse sequential subset selection for supervised video
  summarization.
\newblock In {\em Neural Inform. Process. Syst.}, pages 2069--2077, 2014.

\bibitem{guadarrama2013youtube2text}
S.~Guadarrama, N.~Krishnamoorthy, G.~Malkarnenkar, S.~Venugopalan, R.~Mooney,
  T.~Darrell, and K.~Saenko.
\newblock {Youtube2text}: Recognizing and describing arbitrary activities using
  semantic hierarchies and zero-shot recognition.
\newblock In {\em Proc. Int. Conf. Comput. Vision}, pages 2712--2719, 2013.

\bibitem{gygli2014creating}
M.~Gygli, H.~Grabner, H.~Riemenschneider, and L.~Van~Gool.
\newblock Creating summaries from user videos.
\newblock In {\em European Conf. Comput. Vision}, pages 505--520. Springer,
  2014.

\bibitem{gygli2015video}
M.~Gygli, H.~Grabner, and L.~Van~Gool.
\newblock Video summarization by learning submodular mixtures of objectives.
\newblock In {\em Proc. Conf. Comput. Vision Pattern Recognition}, pages
  3090--3098, 2015.

\bibitem{hassner2013critical}
T.~Hassner.
\newblock A critical review of action recognition benchmarks.
\newblock In {\em Proc. Conf. Comput. Vision Pattern Recognition Workshops},
  pages 245--250, 2013.

\bibitem{hassner2015dense}
T.~Hassner and C.~Liu.
\newblock {\em Dense Image Correspondences for Computer Vision}.
\newblock Springer, 2015.

\bibitem{hochreiter1997long}
S.~Hochreiter and J.~Schmidhuber.
\newblock Long short-term memory.
\newblock {\em Neural computation}, 9(8):1735--1780, 1997.

\bibitem{huang2012multi}
H.~Huang, Y.~Lu, F.~Zhang, and S.~Sun.
\newblock A multi-modal clustering method for web videos.
\newblock In {\em Int. Conf. on Trustworthy Computing and Services}, pages
  163--169. Springer, 2012.

\bibitem{ji20133d}
S.~Ji, W.~Xu, M.~Yang, and K.~Yu.
\newblock 3d convolutional neural networks for human action recognition.
\newblock {\em Trans. Pattern Anal. Mach. Intell.}, 35(1):221--231, 2013.

\bibitem{THUMOS14}
Y.-G. Jiang, J.~Liu, A.~Roshan~Zamir, G.~Toderici, I.~Laptev, M.~Shah, and
  R.~Sukthankar.
\newblock {THUMOS} challenge: Action recognition with a large number of
  classes.
\newblock \url{http://crcv.ucf.edu/THUMOS14/}, 2014.

\bibitem{karpathy2014large}
A.~Karpathy, G.~Toderici, S.~Shetty, T.~Leung, R.~Sukthankar, and L.~Fei-Fei.
\newblock Large-scale video classification with convolutional neural networks.
\newblock In {\em Proc. Conf. Comput. Vision Pattern Recognition}, pages
  1725--1732, 2014.

\bibitem{khosla2013large}
A.~Khosla, R.~Hamid, C.-J. Lin, and N.~Sundaresan.
\newblock Large-scale video summarization using web-image priors.
\newblock In {\em Proc. Conf. Comput. Vision Pattern Recognition}, pages
  2698--2705, 2013.

\bibitem{klein2015associating}
B.~Klein, G.~Lev, G.~Sadeh, and L.~Wolf.
\newblock Associating neural word embeddings with deep image representations
  using fisher vectors.
\newblock In {\em Proc. Conf. Comput. Vision Pattern Recognition}, pages
  4437--4446, 2015.

\bibitem{kliper2012motion}
O.~Kliper-Gross, Y.~Gurovich, T.~Hassner, and L.~Wolf.
\newblock Motion interchange patterns for action recognition in unconstrained
  videos.
\newblock In {\em European Conf. Comput. Vision}, pages 256--269. Springer,
  2012.

\bibitem{kliper2012action}
O.~Kliper-Gross, T.~Hassner, and L.~Wolf.
\newblock The action similarity labeling challenge.
\newblock {\em Trans. Pattern Anal. Mach. Intell.}, 34(3):615--621, 2012.

\bibitem{krishnamoorthy2013generating}
N.~Krishnamoorthy, G.~Malkarnenkar, R.~J. Mooney, K.~Saenko, and S.~Guadarrama.
\newblock Generating natural-language video descriptions using text-mined
  knowledge.
\newblock In {\em AAAI Conf. on Artificial Intelligence}, volume~1, page~2,
  2013.

\bibitem{kuehne2011hmdb}
H.~Kuehne, H.~Jhuang, E.~Garrote, T.~Poggio, and T.~Serre.
\newblock Hmdb: a large video database for human motion recognition.
\newblock In {\em Proc. Int. Conf. Comput. Vision}, pages 2556--2563. IEEE,
  2011.

\bibitem{kulesza2012determinantal}
A.~Kulesza and B.~Taskar.
\newblock Determinantal point processes for machine learning.
\newblock {\em arXiv preprint arXiv:1207.6083}, 2012.

\bibitem{laptev2005space}
I.~Laptev.
\newblock On space-time interest points.
\newblock 64(2-3):107--123, 2005.

\bibitem{lev2015rnn}
G.~Lev, G.~Sadeh, B.~Klein, and L.~Wolf.
\newblock {RNN} fisher vectors for action recognition and image annotation.
\newblock {\em arXiv preprint arXiv:1512.03958}, 2015.

\bibitem{li2016vlad3}
Y.~Li, W.~Li, V.~Mahadevan, and N.~Vasconcelos.
\newblock Vlad3: Encoding dynamics of deep features for action recognition.
\newblock In {\em Proc. Conf. Comput. Vision Pattern Recognition}, pages
  1951--1960, 2016.

\bibitem{lin2004automatic}
C.-Y. Lin and F.~J. Och.
\newblock Automatic evaluation of machine translation quality using longest
  common subsequence and skip-bigram statistics.
\newblock In {\em Proc. Annual Meeting on Association for Computational
  Linguistics}, page 605. Association for Computational Linguistics, 2004.

\bibitem{liu2011sift}
C.~Liu, J.~Yuen, and A.~Torralba.
\newblock {SIFT} flow: Dense correspondence across scenes and its applications.
\newblock {\em Trans. Pattern Anal. Mach. Intell.}, 33(5):978--994, 2011.

\bibitem{mikolov2013efficient}
T.~Mikolov, K.~Chen, G.~Corrado, and J.~Dean.
\newblock Efficient estimation of word representations in vector space.
\newblock {\em arXiv preprint arXiv:1301.3781}, 2013.

\bibitem{oneata2014lear}
D.~Oneata, J.~Verbeek, and C.~Schmid.
\newblock The lear submission at thumos 2014.
\newblock 2014.

\bibitem{over2012trecvid}
P.~Over, G.~Awad, M.~Michel, J.~Fiscus, G.~Sanders, B.~Shaw, A.~F. Smeaton, and
  G.~Qu{\'e}not.
\newblock {TRECVID} 2012--an overview of the goals, tasks, data, evaluation
  mechanisms and metrics.
\newblock In {\em Proceedings of {TRECVID}}, 2012.

\bibitem{owens2016visually}
A.~Owens, P.~Isola, J.~McDermott, A.~Torralba, E.~H. Adelson, and W.~T.
  Freeman.
\newblock Visually indicated sounds.
\newblock In {\em Proc. Conf. Comput. Vision Pattern Recognition}, pages
  2405--2413, 2016.

\bibitem{papineni2002bleu}
K.~Papineni, S.~Roukos, T.~Ward, and W.-J. Zhu.
\newblock {BLEU}: a method for automatic evaluation of machine translation.
\newblock In {\em Proc. annual meeting on association for computational
  linguistics}, pages 311--318. Association for Computational Linguistics,
  2002.

\bibitem{perronnin2010improving}
F.~Perronnin, J.~S{\'a}nchez, and T.~Mensink.
\newblock Improving the fisher kernel for large-scale image classification.
\newblock In {\em European Conf. Comput. Vision}, pages 143--156. Springer,
  2010.

\bibitem{potapov2014category}
D.~Potapov, M.~Douze, Z.~Harchaoui, and C.~Schmid.
\newblock Category-specific video summarization.
\newblock In {\em European Conf. Comput. Vision}, pages 540--555. Springer,
  2014.

\bibitem{rabiner1989tutorial}
L.~R. Rabiner.
\newblock A tutorial on hidden {Markov} models and selected applications in
  speech recognition.
\newblock {\em Proceedings of the IEEE}, 77(2):257--286, 1989.

\bibitem{richard2016temporal}
A.~Richard and J.~Gall.
\newblock Temporal action detection using a statistical language model.
\newblock In {\em Proceedings of the IEEE Conference on Computer Vision and
  Pattern Recognition}, pages 3131--3140, 2016.

\bibitem{rohrbach15cvpr}
A.~Rohrbach, M.~Rohrbach, N.~Tandon, and B.~Schiele.
\newblock A dataset for movie description.
\newblock In {\em Proc. Conf. Comput. Vision Pattern Recognition}, 2015.

\bibitem{Rohrbach2016Large}
A.~Rohrbach, A.~Torabi, T.~Maharaj, M.~Rohrbach, C.~Pal, A.~Courville, and
  B.~Schiele.
\newblock The large scale movie description and understanding challenge {(LSMDC
  2016)}, howpublished = {Available:~\url{http://tinyurl.com/zabh4et}}, month =
  {September}, year = {2016}.

\bibitem{lsmdc2015}
A.~Rohrbach, A.~Torabi, M.~Rohrbach, N.~Tandon, P.~Chris, L.~Hugo, C.~Aaron,
  and B.~Schiele.
\newblock Movie description.
\newblock {\em arXiv preprint}, 2016.

\bibitem{russakovsky2015imagenet}
O.~Russakovsky, J.~Deng, H.~Su, J.~Krause, S.~Satheesh, S.~Ma, Z.~Huang,
  A.~Karpathy, A.~Khosla, M.~Bernstein, et~al.
\newblock Imagenet large scale visual recognition challenge.
\newblock {\em Int. J. Comput. Vision}, 115(3):211--252, 2015.

\bibitem{shetty2015video}
R.~Shetty and J.~Laaksonen.
\newblock Video captioning with recurrent networks based on frame-and
  video-level features and visual content classification.
\newblock {\em arXiv preprint arXiv:1512.02949}, 2015.

\bibitem{shou2016temporal}
Z.~Shou, D.~Wang, and S.-F. Chang.
\newblock Temporal action localization in untrimmed videos via multi-stage
  cnns.
\newblock In {\em Proceedings of the IEEE Conference on Computer Vision and
  Pattern Recognition}, pages 1049--1058, 2016.

\bibitem{simonyan2013deep}
K.~Simonyan, A.~Vedaldi, and A.~Zisserman.
\newblock Deep fisher networks for large-scale image classification.
\newblock In {\em Neural Inform. Process. Syst.}, pages 163--171, 2013.

\bibitem{simonyan2014two}
K.~Simonyan and A.~Zisserman.
\newblock Two-stream convolutional networks for action recognition in videos.
\newblock In {\em Neural Inform. Process. Syst.}, pages 568--576, 2014.

\bibitem{simonyan2014very}
K.~Simonyan and A.~Zisserman.
\newblock Very deep convolutional networks for large-scale image recognition.
\newblock {\em arXiv preprint arXiv:1409.1556}, 2014.

\bibitem{song2015tvsum}
Y.~Song, J.~Vallmitjana, A.~Stent, and A.~Jaimes.
\newblock Tvsum: Summarizing web videos using titles.
\newblock In {\em Proc. Conf. Comput. Vision Pattern Recognition}, pages
  5179--5187, 2015.

\bibitem{soomro2012ucf101}
K.~Soomro, A.~R. Zamir, and M.~Shah.
\newblock Ucf101: A dataset of 101 human actions classes from videos in the
  wild.
\newblock {\em arXiv preprint arXiv:1212.0402}, 2012.

\bibitem{srivastava2015unsupervised}
N.~Srivastava, E.~Mansimov, and R.~Salakhutdinov.
\newblock Unsupervised learning of video representations using {LSTM}s.
\newblock In {\em Int. Conf. Mach. Learning}, volume~2, 2015.

\bibitem{sydorov2014deep}
V.~Sydorov, M.~Sakurada, and C.~H. Lampert.
\newblock Deep fisher kernels-end to end learning of the fisher kernel {GMM}
  parameters.
\newblock In {\em Proc. Conf. Comput. Vision Pattern Recognition}, pages
  1402--1409, 2014.

\bibitem{taylor2010convolutional}
G.~W. Taylor, R.~Fergus, Y.~LeCun, and C.~Bregler.
\newblock Convolutional learning of spatio-temporal features.
\newblock In {\em European Conf. Comput. Vision}, pages 140--153. Springer,
  2010.

\bibitem{thomason2014integrating}
J.~Thomason, S.~Venugopalan, S.~Guadarrama, K.~Saenko, and R.~J. Mooney.
\newblock Integrating language and vision to generate natural language
  descriptions of videos in the wild.
\newblock In {\em COLING}, volume~2, page~9, 2014.

\bibitem{AtorabiMVAD2015}
A.~Torabi, C.~Pal, H.~Larochelle, and A.~Courville.
\newblock Using descriptive video services to create a large data source for
  video annotation research.
\newblock {\em arXiv preprint}, 2015.

\bibitem{tran2015learning}
D.~Tran, L.~Bourdev, R.~Fergus, L.~Torresani, and M.~Paluri.
\newblock Learning spatiotemporal features with 3d convolutional networks.
\newblock In {\em Proc. Int. Conf. Comput. Vision}, pages 4489--4497, 2015.

\bibitem{vedantam2015cider}
R.~Vedantam, C.~Lawrence~Zitnick, and D.~Parikh.
\newblock Cider: Consensus-based image description evaluation.
\newblock In {\em Proc. Conf. Comput. Vision Pattern Recognition}, pages
  4566--4575, 2015.

\bibitem{venugopalan2015sequence}
S.~Venugopalan, M.~Rohrbach, J.~Donahue, R.~Mooney, T.~Darrell, and K.~Saenko.
\newblock Sequence to sequence-video to text.
\newblock In {\em Proc. Conf. Comput. Vision Pattern Recognition}, pages
  4534--4542, 2015.

\bibitem{regularized_cca}
H.~D. Vinod.
\newblock {Canonical ridge and econometrics of joint production}.
\newblock {\em Journal of Econometrics}, 4(2):147--166, May 1976.

\bibitem{vinyals2015show}
O.~Vinyals, A.~Toshev, S.~Bengio, and D.~Erhan.
\newblock Show and tell: A neural image caption generator.
\newblock In {\em Proc. Conf. Comput. Vision Pattern Recognition}, pages
  3156--3164, 2015.

\bibitem{wang2013action}
H.~Wang and C.~Schmid.
\newblock Action recognition with improved trajectories.
\newblock In {\em Proc. Int. Conf. Comput. Vision}, pages 3551--3558, 2013.

\bibitem{wang2014action}
L.~Wang, Y.~Qiao, and X.~Tang.
\newblock Action recognition and detection by combining motion and appearance
  features.
\newblock {\em THUMOS14 Action Recognition Challenge}, 1:2, 2014.

\bibitem{wang2015action}
L.~Wang, Y.~Qiao, and X.~Tang.
\newblock Action recognition with trajectory-pooled deep-convolutional
  descriptors.
\newblock In {\em Proc. Conf. Comput. Vision Pattern Recognition}, pages
  4305--4314, 2015.

\bibitem{wei2010multimodal}
S.~Wei, Y.~Zhao, Z.~Zhu, and N.~Liu.
\newblock Multimodal fusion for video search reranking.
\newblock {\em Trans. on Knowledge and Data Engineering}, 22(8):1191--1199,
  2010.

\bibitem{yao2015describing}
L.~Yao, A.~Torabi, K.~Cho, N.~Ballas, C.~Pal, H.~Larochelle, and A.~Courville.
\newblock Describing videos by exploiting temporal structure.
\newblock In {\em Proc. Int. Conf. Comput. Vision}, pages 4507--4515, 2015.

\bibitem{yeung2016end}
S.~Yeung, O.~Russakovsky, G.~Mori, and L.~Fei-Fei.
\newblock End-to-end learning of action detection from frame glimpses in
  videos.
\newblock In {\em Proceedings of the IEEE Conference on Computer Vision and
  Pattern Recognition}, pages 2678--2687, 2016.

\bibitem{yu2016video}
Y.~Yu, H.~Ko, J.~Choi, and G.~Kim.
\newblock Video captioning and retrieval models with semantic attention.
\newblock {\em arXiv preprint arXiv:1610.02947}, 2016.

\bibitem{yuan2016temporal}
J.~Yuan, B.~Ni, X.~Yang, and A.~A. Kassim.
\newblock Temporal action localization with pyramid of score distribution
  features.
\newblock In {\em Proc. Conf. Comput. Vision Pattern Recognition}, pages
  3093--3102, 2016.

\bibitem{zhang1997integrated}
H.~J. Zhang, J.~Wu, D.~Zhong, and S.~W. Smoliar.
\newblock An integrated system for content-based video retrieval and browsing.
\newblock {\em Pattern recognition}, 30(4):643--658, 1997.

\bibitem{zhang2016summary}
K.~Zhang, W.-L. Chao, F.~Sha, and K.~Grauman.
\newblock Summary transfer: Exemplar-based subset selection for video
  summarizatio.
\newblock In {\em Proc. Conf. Comput. Vision Pattern Recognition}, 2016.

\bibitem{zhang2016video}
K.~Zhang, W.-L. Chao, F.~Sha, and K.~Grauman.
\newblock Video summarization with long short-term memory.
\newblock In {\em European Conf. Comput. Vision}, 2016.

\bibitem{zhao2014quasi}
B.~Zhao and E.~P. Xing.
\newblock Quasi real-time summarization for consumer videos.
\newblock In {\em Proc. Conf. Comput. Vision Pattern Recognition}, pages
  2513--2520, 2014.

\end{thebibliography}
}

\end{document}